\newif\ifisTR

\isTRtrue

\documentclass{article} 

\usepackage{fullpage}

\usepackage{microtype}
\usepackage{graphicx}
\usepackage{booktabs} 
\usepackage{placeins}
\usepackage{algorithm}
\usepackage{algorithmic}
\usepackage{subcaption}
\usepackage{tcolorbox}
\usepackage{nicefrac}  
\usepackage{enumitem}
\usepackage{multirow}
\usepackage{colortbl}
\usepackage{xspace}
\usepackage{wrapfig}
\usepackage{amsmath}
\usepackage{amssymb}
\usepackage{mathtools}
\usepackage[export]{adjustbox}
\usepackage{amsthm}
\usepackage{xcolor}

\definecolor{CColor}{rgb}{0.01,0.31,0.59}
\definecolor{GGray}{rgb}{0.80,0.90,1}
\definecolor{Shady}{rgb}{0.9,0.9,0.9}
\definecolor{kaistblue}{RGB}{20,135,200}
\definecolor{kaistdarkblue}{RGB}{0,65,145}
\definecolor{urbanablue}{RGB}{19,41,75}
\definecolor{urbanaorange}{RGB}{232,74,39}
\definecolor{drp}{rgb}{0.53,0.15,0.34}
\usepackage[plainpages=false,pdfpagelabels,colorlinks=true,linkcolor=CColor,citecolor=CColor,urlcolor=CColor]{hyperref}

\usepackage{tikz}

\usepackage[capitalize,noabbrev]{cleveref}

\usepackage{overpic}

\theoremstyle{plain}

\theoremstyle{definition}

\theoremstyle{remark}

\definecolor{mygray}{gray}{0.85}
\definecolor{LightBlue}{cmyk}{0.06, 0.03, 0.01, 0.0}

\usepackage[textsize=tiny]{todonotes}

\usepackage[round,semicolon,sort]{natbib}

\renewcommand{\cite}[1]{\citep{#1}}

\title{Demystifying the Roles of LLM Layers in Retrieval, Knowledge, and Reasoning}
\date{}
\author{
 Xinyuan Song\textsuperscript{1},
 Keyu Wang\textsuperscript{2},
 Pengxiang Li\textsuperscript{3},
 Lu Yin\textsuperscript{4},
 Shiwei Liu\textsuperscript{5,6}
\\
 \textsuperscript{1}Emory University, USA\\
 \textsuperscript{2}University of Tuebingen, Germany \\
 \textsuperscript{3}Hong Kong Polytechic University, Hong Kong \\
 \textsuperscript{4}University of Surrey, UK \\
 \textsuperscript{5}Max Planck Institute for Intelligent Systems,\\
 \textsuperscript{6}ELLIS Institute Tübingen, \\
}
\begin{document}
%

\maketitle
\begin{abstract}
Recent studies suggest that the deeper layers of Large Language Models (LLMs) contribute little to representation learning and can often be removed without significant performance loss. However, such claims are typically drawn from narrow evaluations and may overlook important aspects of model behavior. In this work, we present a systematic study of depth utilization across diverse dimensions, including evaluation protocols, task categories, and model architectures. Our analysis confirms that very deep layers are generally less effective than earlier ones, but their contributions vary substantially with the evaluation setting. Under likelihood-based metrics without generation, pruning most layers preserves performance, with only the initial few being critical. By contrast, generation-based evaluation uncovers indispensable roles for middle and deeper layers in enabling reasoning and maintaining long-range coherence. We further find that knowledge and retrieval are concentrated in shallow components, whereas reasoning accuracy relies heavily on deeper layers—yet can be reshaped through distillation. These results highlight that depth usage in LLMs is highly heterogeneous and context-dependent, underscoring the need for task-, metric-, and model-aware perspectives in both interpreting and compressing large models. Our code is avalailable at \url{https://github.com/Hik289/llm-layer-importance.git}.
\end{abstract}

\section{Introduction}
\label{sec:intro}

Deep neural networks have long exhibited depth-related challenges~\cite{hanin2023randomfullyconnectedneural}, including vanishing or exploding gradients~\cite{hochreiter1991untersuchungen}, rank collapse~\cite{bhojanapalli2020low}, and representational redundancy~\cite{raghu2017svcca}. These effects reduce the marginal contribution of later layers during training and inference, even when optimization succeeds at lowering the loss~\cite{dong2021attention}. Large language models (LLMs) amplify these issues by pushing depth and width to the extreme: they contain billions of parameters and tens to hundreds of Transformer blocks~\cite{brown2020language}. Such scale increases sensitivity to variance growth~\cite{takase2025spikemorestabilizingpretraining} and weak inter-layer signal propagation~\cite{dong2021attention}, making the effectiveness of very deep layers a central concern.  A recent study, \textbf{Curse of Depth}~\cite{sun2025cursedepthlargelanguage}, reports that deeper layers in modern LLMs can become ineffective due to variance explosion. Other analyses indicate that deeper layers may suffer from rank collapse or excessive feature overlap, limiting their ability to produce new, useful representations~\cite{li2025mixlnunleashingpowerdeeper,gromov2024unreasonable,men2024shortgpt}. 

This phenomenon presents both opportunities and challenges. On the opportunity side, deeper layers can be compressed more aggressively~\cite{lu2024alphapruning,dumitru2024changeconstantdynamicllm}, and in some cases entire deep layers can be pruned without compromising performance~\cite{muralidharan2024compactlanguagemodelspruning,siddiqui2024deeperlookdepthpruning,nepal2025layerimportancemathematicalreasoning}. However, such claims are often based on narrow evaluations that may overlook important aspects of model behavior. Conventional benchmarks typically emphasize a limited set of tasks or metrics, which fail to capture the full range of capabilities that deeper layers might provide~\cite{skean2025layerlayeruncoveringhidden,srivastava2022beyond}. In more complex scenarios—such as knowledge-intensive reasoning~\cite{petroni2021kilt}, mathematical problem solving~\cite{srivastava2022beyond,cobbe2021training}, or retrieval-augmented tasks~\cite{lewis2020retrieval}—these layers may play a much more significant role. In light of these concerns, a key challenge remains unresolved: \textit{How can we rigorously evaluate the contribution of each layer to overall LLM performance under different evaluation protocols?}

Building on these concerns, we present a systematic study of depth utilization across evaluation protocols, task categories, and model architectures. We first examine how different \textbf{evaluation protocols} affect conclusions about layer-wise importance, considering three protocols: \emph{log-likelihood default}~\cite{hendrycks2021mmlu}, \emph{log-likelihood continuation}~\cite{paperno2016lambada}, and \emph{generation until}~\cite{liang2022helm,eval-harness}. For each, we prune layers and measure performance degradation to quantify their contribution. Our analysis yields three \textbf{key findings}: (1) the perceived importance of layers is highly heterogeneous across evaluation metrics, underscoring the need for more principled and task-aware evaluation methodologies. (2) likelihood-based metrics emphasize shallow representations, with degradation concentrated in early layers; (3) generation exposes vulnerabilities in middle and deep layers, highlighting their role in reasoning and coherence; and 

We next examine \textbf{knowledge and retrieval tasks}~\cite{zellers2019hellaswag,amini2019mathqa} to probe how different functional demands shape layer usage. These tasks range from commonsense reasoning, which emphasizes shallow contextual plausibility, to retrieval-oriented settings~\cite{mihaylov2018openbookqa,gu2023kvretrieval} that require accessing stored or external information. Through layer pruning, we identify three \textbf{key findings}: (1) shallow layers are critical, with their removal causing sharp degradation while deeper layers contribute less directly; (2) retrieval augmented with external evidence extends robustness into middle and deeper layers; and (3) knowledge and retrieval abilities are not uniformly distributed but localized to specific layers and even individual heads, highlighting both opportunities for targeted compression and challenges in preserving task fidelity.

Finally, we study \textbf{reasoning tasks and distilled models}~\cite{rein2023gpqa,aime,math500}. On benchmarks such as GSM8k~\cite{cobbe2021gsm8k}, reasoning accuracy depends strongly on middle and deep layers, with specific reasoning heads identified as critical. Distillation not only improves baseline accuracy but also redistributes reasoning capacity more evenly across depth, thereby enhancing robustness under ablation. These findings highlight how reasoning differs from knowledge and retrieval tasks and how distillation reshapes depth dependence in LLMs.

Our extensive controlled experiments across diverse tasks, evaluation metrics, and model families consistently show that the contribution of layers is highly non-uniform. Different tasks activate different depths, different metrics emphasize different subsets of layers, and different model designs further modulate these effects. These findings demonstrate that understanding depth usage in LLMs requires a \textbf{task-, metric-, and model-aware perspective}, which is essential to avoid experimental bias and to ensure that conclusions about layer importance are both reliable and broadly applicable.

\section{Evaluation Protocol Matters}\label{sec:eval}  

In this section, we examine how various evaluation protocols influence the assessment of layer pruning. Our goal is to understand whether likelihood-based metrics and generation-based metrics reveal consistent or divergent patterns of layer importance.  

We evaluate the impact of different evaluation metrics under layer pruning using two representative LLMs, \textbf{LLaMA-3.1-8B}~\cite{touvron2023llama} and \textbf{Qwen3-8B}~\cite{bai2023qwen}. All experiments are conducted on the \textbf{MMLU benchmark}~\cite{hendrycks2021mmlu}, a widely adopted testbed for general knowledge and reasoning. We consider \textbf{log-likelihood default}, which scores multiple-choice answers based on option likelihood; \textbf{log-likelihood continuation}, which evaluates token-level continuation probability; and \textbf{generation until}, which measures open-ended autoregressive generation until the final answer. For each setting, we systematically ablate layers and measure performance degradation as a function of layer index, reporting both accuracy ($\mu$) and relative change ($\Delta \mu$, defined as the difference between the full model and the layer-pruned model).

\begin{figure}[!ht]
    \centering
    \includegraphics[width=\linewidth]{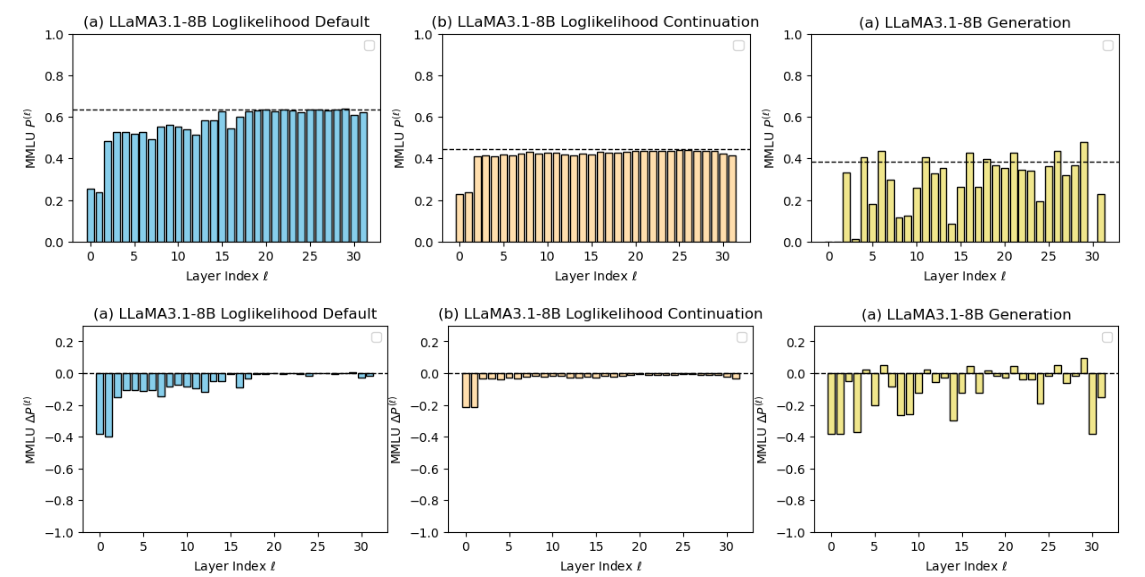}
    \caption{Layer pruning results of \textbf{LLaMA-3.1-8B} on the \textbf{MMLU} benchmark under three evaluation protocols: log-likelihood default (left), log-likelihood continuation (middle), and generation until (right). We report accuracy ($\mu$) and relative change ($\Delta \mu$) across layers.}
    \label{fig:llama_mmlu}
        
\end{figure}

\begin{figure}[!ht]
    \centering
    \includegraphics[width=\linewidth]{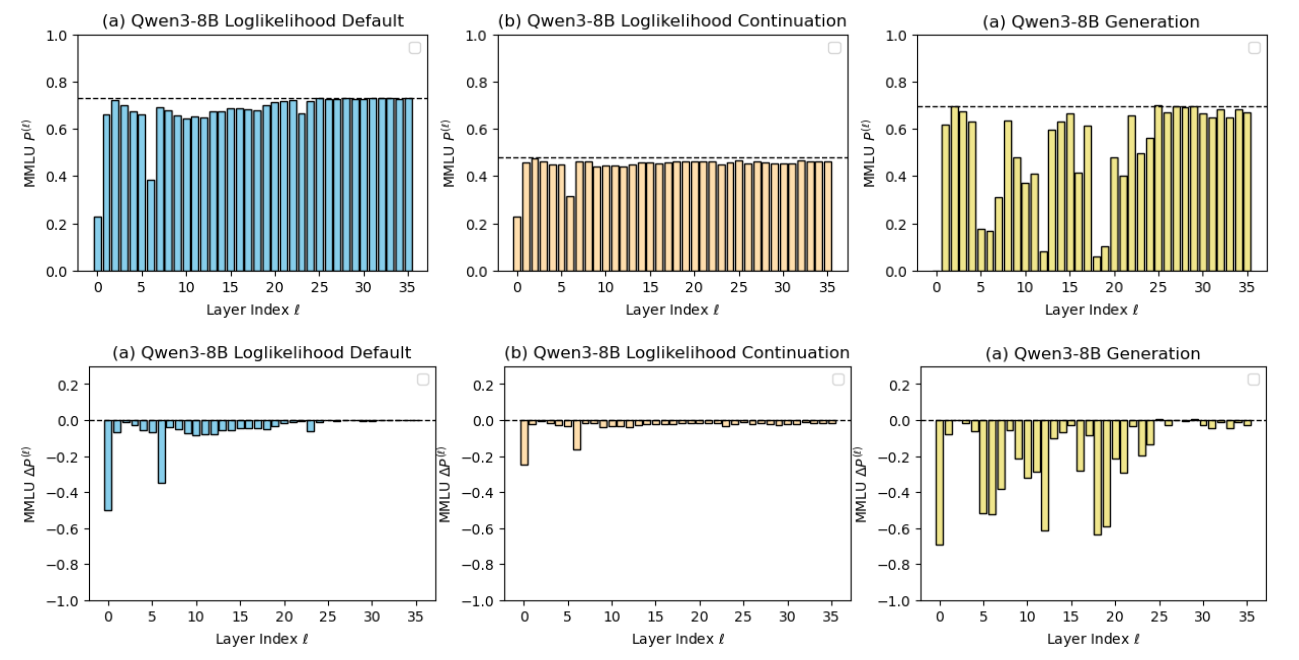}
    \caption{Layer pruning results of \textbf{Qwen3-8B} on the \textbf{MMLU} benchmark under the same three evaluation protocols. Accuracy ($\mu$) and relative change ($\Delta \mu$) are shown across layer indices.}
    \label{fig:qwen_mmlu}
\end{figure}

Results in Figures~\ref{fig:llama_mmlu} and~\ref{fig:qwen_mmlu} show that in log-likelihood–based evaluations (multiple-choice or continuation), degradation is concentrated in the earliest layers, suggesting reliance on shallow representations to maintain token-level coherence. In contrast, the generation-based evaluation reveals substantial drops across middle and deeper layers, indicating that multi-step reasoning and long-range consistency require contributions from the entire depth of the network. Together, these results demonstrate that likelihood-based evaluations substantially underestimate the fragility of compressed models, whereas generation more faithfully captures the dependence of LLMs on hierarchical depth.

\section{Layer Importance in Knowledge Tasks}

In this section, we analyze how layer pruning affects performance on knowledge-intensive tasks. We focus on commonsense reasoning benchmarks to identify whether shallow or deep layers contribute more critically to accuracy.  

\subsection{commonsense dataset}
\begin{figure}[!ht]
    
    \centering
        \resizebox{0.8\columnwidth}{!}{
    \includegraphics[width=\linewidth]{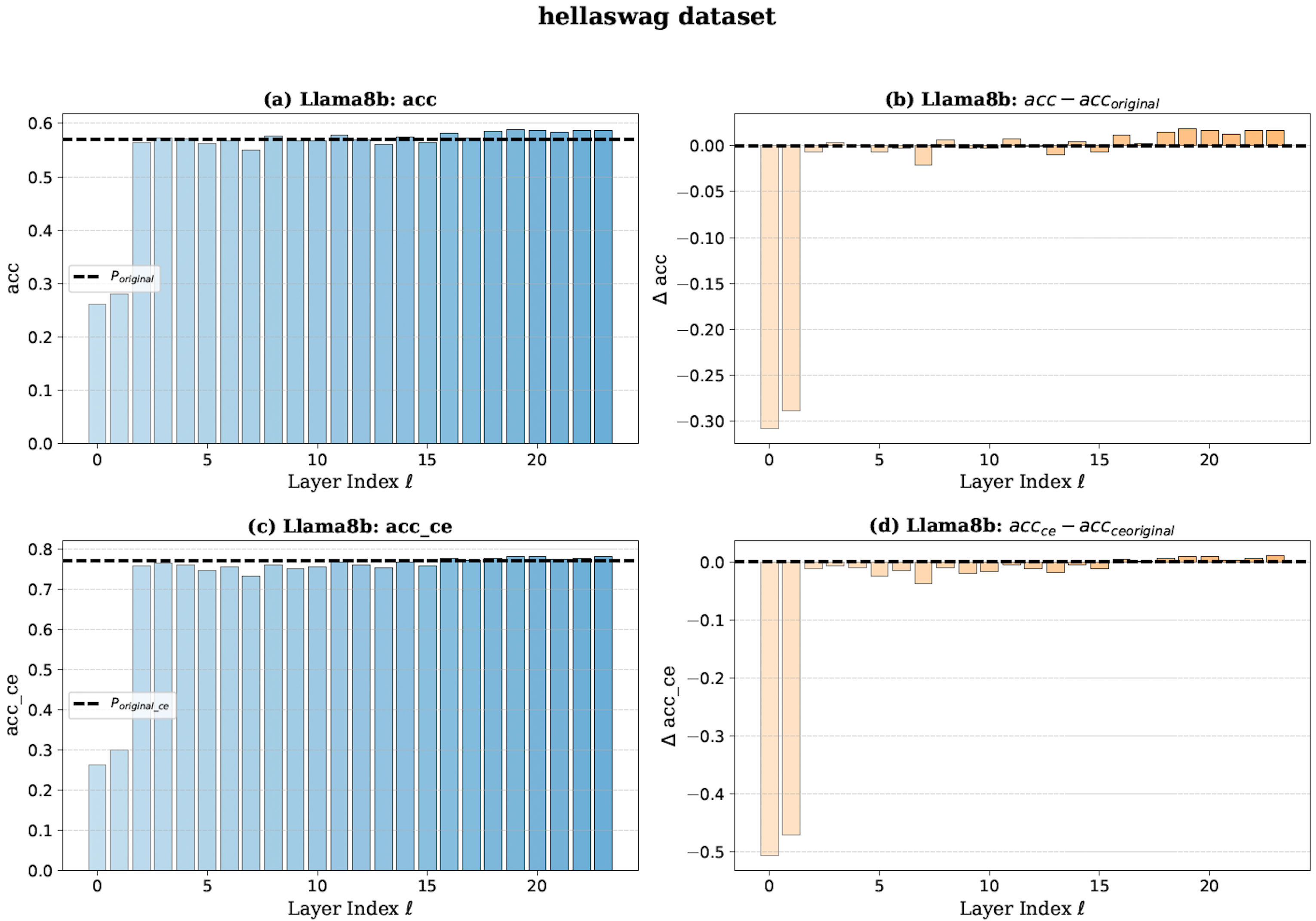}}
    \caption{layer pruning results of \textbf{LLaMA-3.1-8B} on the \textbf{HellaSwag} dataset. We report both standard accuracy ($\textbf{acc}$) and cross-entropy–based accuracy (\textbf{acc\_ce}), along with their relative differences compared to the unablated model across layers.}
    \label{fig:hellaswag_ablation}
    
\end{figure}

We evaluate \textbf{LLaMA-3.1-8B} on the \textbf{HellaSwag}~\cite{zellers2019hellaswag} commonsense dataset under a \textbf{layer pruning} setting. Three evaluation metrics are considered: the standard multiple-choice accuracy (\textbf{acc}), a cross-entropy–based accuracy (\textbf{acc\_ce}), based on the \textbf{log-likelihood default} evaluation protocol as described in Section~\ref{sec:eval}. For each metric, we report both the absolute performance and the relative difference compared to the unablated model as functions of layer index. The results are summarized in Figure~\ref{fig:hellaswag_ablation}.

As shown in the figure, ablating early layers leads to substantial performance degradation, with accuracy drops up to $-0.3$ in \textbf{acc} and $-0.5$ in \textbf{acc\_ce}. In contrast, middle and later layers exhibit negligible changes, and in some cases even slight improvements. These findings suggest that commonsense continuation tasks such as HellaSwag rely heavily on shallow representations, while deeper layers play a less critical role in maintaining accuracy.

\subsection{Math Problem Solving}

We also evaluate \textbf{LLaMA-3.1-8B} on the \textbf{MathQA} dataset~\cite{amini2019mathqa}, a large-scale benchmark for mathematical problem solving. Our goal is to examine whether reasoning in math benchmarks exhibits broader or different sensitivity to layer pruning compared to commonsense tasks.  

\begin{figure}[!ht]
    \centering
    \includegraphics[width=0.8\linewidth]{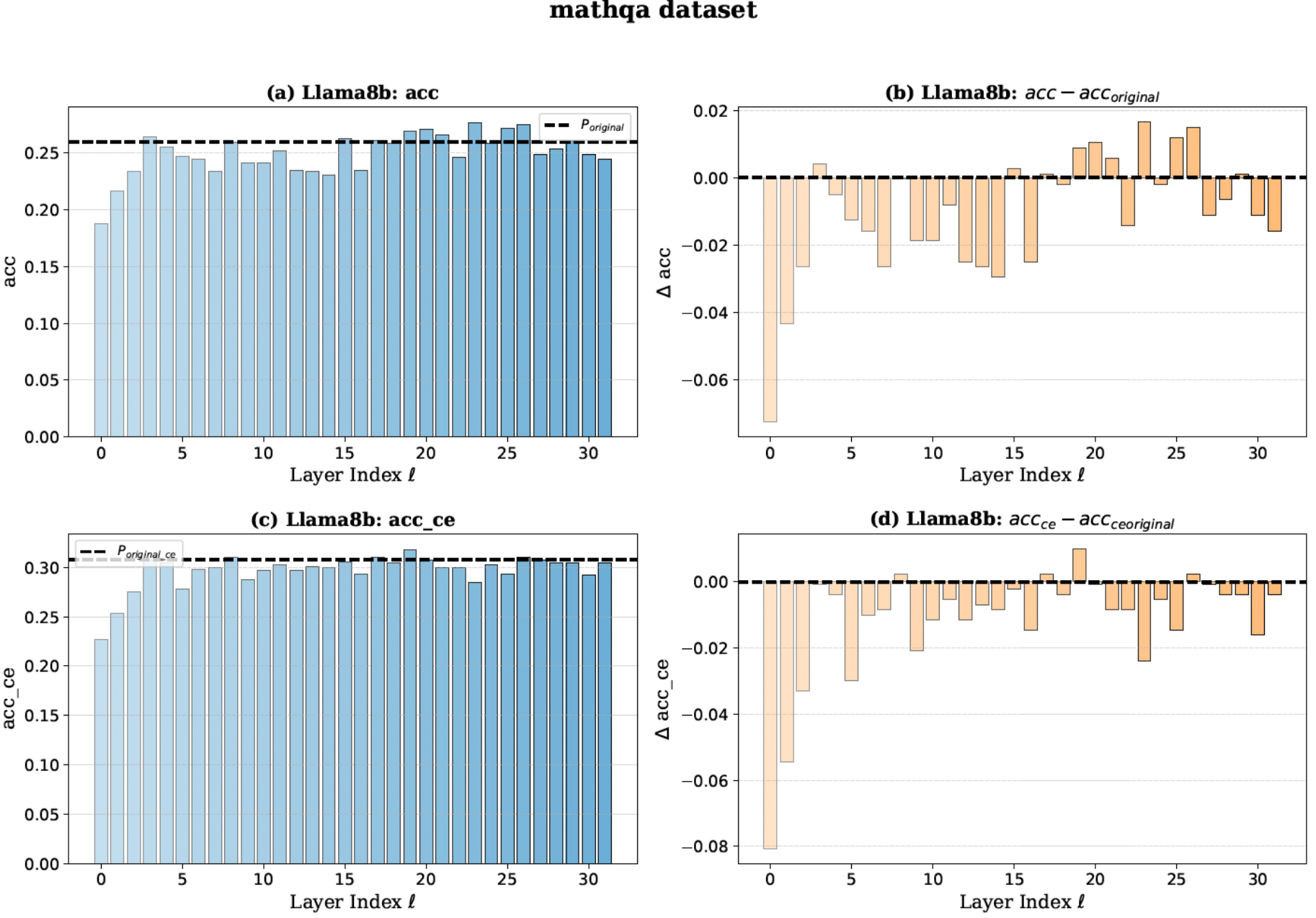}
    \caption{layer pruning results of \textbf{LLaMA-3.1-8B} on the \textbf{MathQA} dataset. }
    \label{fig:mathqa_ablation}
    
\end{figure}

As in the HellaSwag evaluation, we report both the standard multiple-choice accuracy (\textbf{acc}) and the cross-entropy–based accuracy (\textbf{acc\_ce}), analyzing their absolute values and relative deviations with respect to the unablated model. The results are shown in Figure~\ref{fig:mathqa_ablation}.

In contrast to \textbf{HellaSwag}, \textbf{MathQA} exhibits broader sensitivity to layer pruning across the network. Removing early layers leads to moderate drops, with reductions up to $-0.06$ in \textbf{acc} and $-0.08$ in \textbf{acc\_ce}, while degradations also persist into the middle layers. This pattern indicates that mathematical reasoning tasks require both shallow and intermediate representations, reflecting a cumulative reliance on symbolic manipulation and semantic integration distributed across model depth.

\section{Layer Importance in Retrieval Tasks}

A key question in understanding large language models is how retrieval ability is distributed across depth. To investigate this, we evaluate the effect of layer pruning on retrieval-oriented tasks.

\subsection{KV retrieval}

We evaluate \textbf{LLaMA-3.1-8B} on the \textbf{KV Retrieval task}~\cite{bick2025understandingskillgaprecurrent} under layer pruning. The task requires retrieving the correct key–value pair from stored memory. We report accuracy ($\mu$) and relative change ($\Delta \mu$), based on the \textbf{log-likelihood default} evaluation protocol across layers. 
\begin{figure}[!ht]
    
    \centering
            \resizebox{0.8\columnwidth}{!}{
    \includegraphics[width=\linewidth]{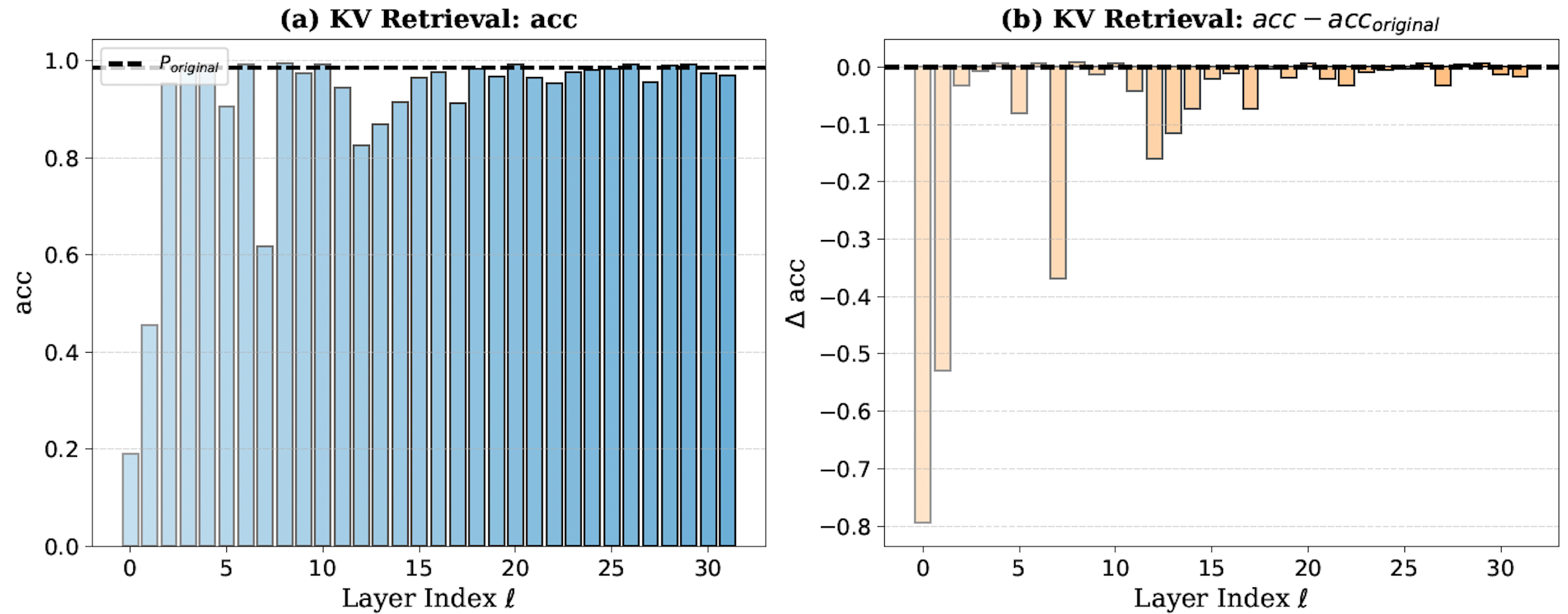}}
    \caption{layer pruning results of \textbf{LLaMA-3.1-8B} on the \textbf{KV Retrieval task}. (a) accuracy $\mu$ (blue), (b) $\Delta \mu$ (yellow).}
    \label{fig:kv_ablation}
    
\end{figure}

See Figure~\ref{fig:kv_ablation}. Results show that shallow layers are critical for retrieval, as ablations in the first few layers cause sharp accuracy drops (up to $-0.8$). Beyond the lower layers, the model maintains near-perfect retrieval performance, and $\Delta \mu$ remains close to zero. This indicates that retrieval ability is encoded primarily in early representations, while middle and deeper layers contribute less to this task. 

\subsection{Retrieval Augmentation}\label{retrieval_aug}

\begin{figure}[!ht]
    \centering
    \includegraphics[width=\linewidth]{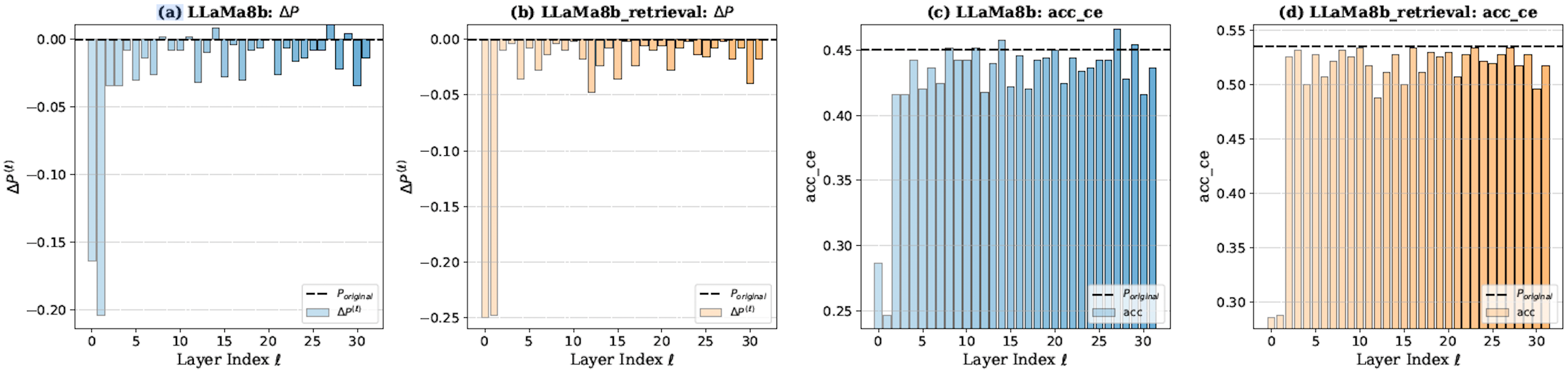}
    \caption{Layer pruning results of \textbf{LLaMA-3.1-8B} on the \textbf{OpenBookQA} benchmark under non-retrieval and retrieval-augmented settings. Blue curves denote standard accuracy (\textbf{acc} $\mu$ and $\Delta \mu$), while yellow curves denote cross-entropy–based accuracy (\textbf{acc\_ce}).}

    \label{fig:retrieval_aug}
    
\end{figure}

Building on the analysis of \textbf{KV Retrieval}, which highlights the role of shallow layers in memorization-based retrieval, we next examine how retrieval compares to non-retrieval settings in question answering tasks. Specifically, we investigate the effect of retrieval augmentation on \textbf{LLaMA-3.1-8B} by performing layer pruning experiments on the \textbf{OpenBookQA} and \textbf{CloseBookQA} benchmarks~\cite{mihaylov2018openbookqa} (Figure~\ref{fig:retrieval_aug}). In this setting, \textbf{OpenBookQA} serves as the retrieval-augmented mode, where the model incorporates external evidence before answering, while \textbf{CloseBookQA} serves as the non-retrieval baseline, where the model answers questions directly without access to retrieval.We report both standard accuracy (\textbf{acc}) and cross-entropy–based accuracy (\textbf{acc\_ce}) as evaluation metrics across layers.

Results show that retrieval consistently improves robustness across almost all layers, with the largest benefits appearing in middle and deeper layers. While ablation degrades both settings, the retrieval-augmented model maintains higher accuracy and exhibits smaller performance drops. This indicates that retrieval not only boosts baseline accuracy but also enhances stability against layer pruning.

\subsection{LLaMA-1 Results}\label{llama1}

Since our experiments on \textbf{LLaMA-3.1-8B} did not reveal a clear dependence of retrieval tasks on intermediate layers, we hypothesize that the strength of the larger model may obscure such effects. To further probe this question, we turn to a simpler model, \textbf{LLaMA-1 7B}~\cite{touvron2023llama}, and perform layer pruning on the KV Retrieval task using the same  evaluation metrics—accuracy ($\mu$) and relative change ($\Delta \mu$) based on the \textbf{log-likelihood default} evaluation protocol.

\begin{figure}[!ht]
    
    \centering
            \resizebox{0.8\columnwidth}{!}{
    \includegraphics[width=\linewidth]{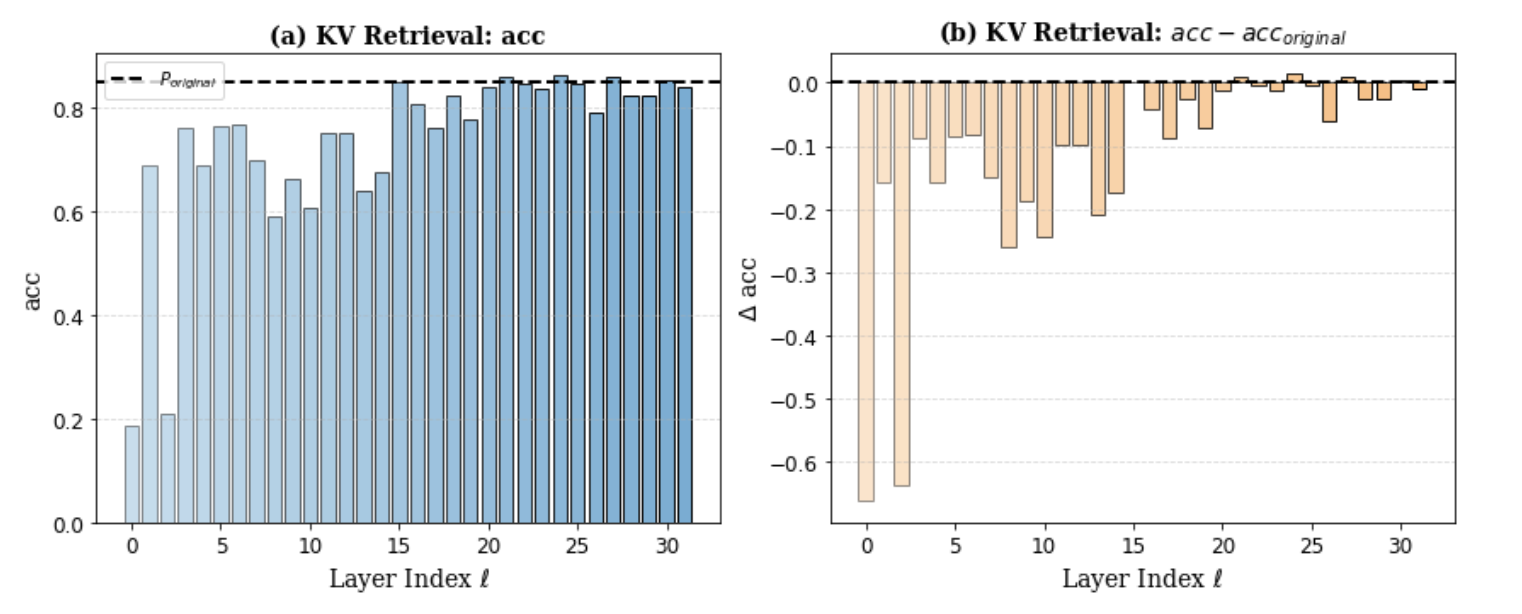}}
    \caption{\textbf{LLaMA-1 7B} on the KV Retrieval task under layer pruning. (a) accuracy $\mu$ (blue), (b) $\Delta \mu$ (yellow).}
    \label{fig:llama1_kv}
    
\end{figure}

For \textbf{KV Retrieval} (Figure~\ref{fig:llama1_kv}), performance remains near-perfect once the lower layers are preserved, but pruning within the first two layers leads to sharp accuracy drops (up to $-0.6$). Unlike the results observed on \textbf{LLaMA-3.1-8B}, where accuracy stabilizes quickly after shallow layers, the \textbf{LLaMA-1 7B} model also shows noticeable degradations in some middle layers. This contrast suggests that the impact of layer pruning on retrieval is model-dependent, and smaller models may distribute retrieval capacity less compactly across depth compared to larger models.

\section{Retrieval Head}\label{retrieval_head}

To further probe the phenomenon observed in Section~\ref{llama1} and Figure~\ref{fig:llama1_kv}, where \textbf{LLaMA-1 7B} shows pronounced accuracy drops at layers 8 and 10 on the \textbf{KV Retrieval task}, we conduct head pruning on these two layers. The goal is to localize the source of degradation and identify specific attention heads that dominate retrieval performance.

\begin{figure}[!ht]
    \centering
    \includegraphics[width=0.8\linewidth]{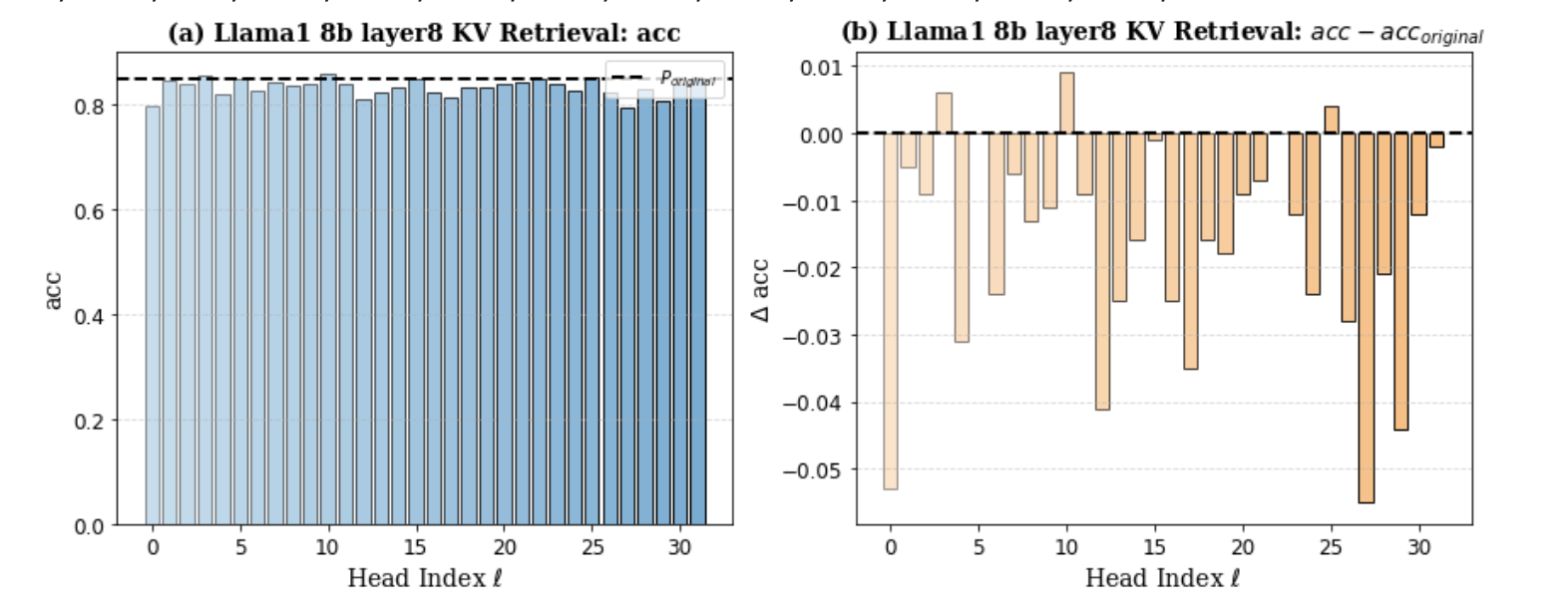}
    \caption{LLaMA-1 7B \textbf{head pruning at layer 8} on KV Retrieval: (a) accuracy per head, (b) $\Delta \mu$ per head.}
    \label{fig:kv_head8}
\end{figure}

\begin{figure}[!ht]
    \centering
    \includegraphics[width=0.8\linewidth]{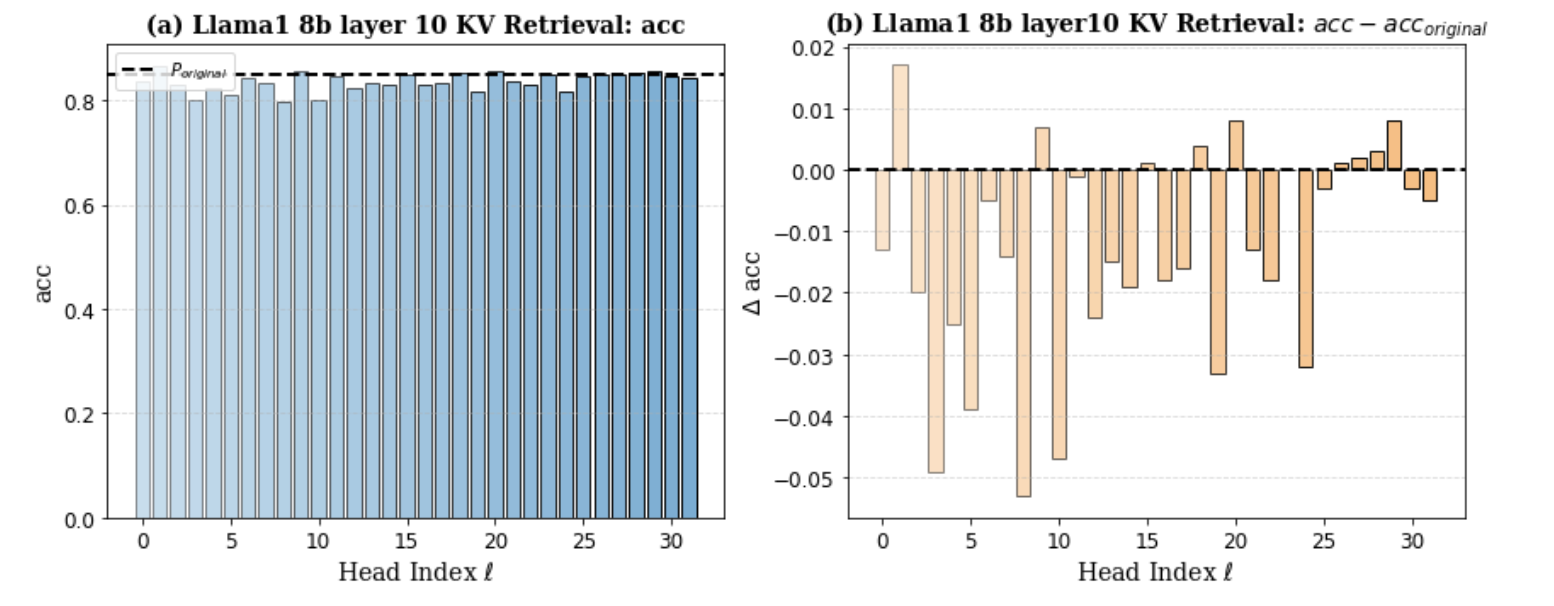}
    \caption{LLaMA-1 7B \textbf{head pruning at layer 10} on KV Retrieval: (a) accuracy per head, (b) $\Delta \mu$ per head.}
    \label{fig:kv_head10}
\end{figure}

Head pruning experiments reveal that retrieval ability is concentrated in specific attention heads rather than being uniformly distributed. At layer 8, several heads prove to be critical, with individual ablations causing notable performance drops ($\Delta \mu$ up to $-0.05$), indicating that retrieval is strongly tied to particular heads in this shallow layer (Figure~\ref{fig:kv_head8}). At layer 10, a similar but weaker pattern emerges, as some heads disproportionately affect performance, though the magnitude of degradation is smaller compared to layer 8 (Figure~\ref{fig:kv_head10}). These results highlight that retrieval depends on a sparse set of specialized heads in shallow and mid-level layers, making their identification crucial for both understanding retrieval mechanisms and maintaining performance under pruning or compression.

\section{Layer Importance in Reasoning Tasks}

Previous sections examined the role of depth in knowledge and retrieval tasks, here we investigate whether a distinct reasoning layer emerges and how its position within the network affects performance. 

\subsection{QWEN Result}

We conduct \textbf{layer pruning} experiments on the \textbf{GSM8K 8-shot}~\cite{cobbe2021gsm8k} benchmark using four models: \textbf{Qwen3-8B}~\cite{bai2023qwen}, \textbf{Qwen3-8B without thinking mode}, and \textbf{LLaMA-3.1-8B}~\cite{touvron2023llama}. 
We selected this range of models to capture varying levels of chain-of-thought (CoT)~\cite{wei2022chain} capability. Qwen3-8B generally exhibit stronger CoT performance, whereas LLaMA-3.1-8B and the Qwen3-8B variant with disabled thinking mode display comparatively weaker CoT. For each model, we measure performance degradation across layers, reporting accuracy ($\mu$) and relative accuracy change ($\Delta \mu$) based on the \textbf{generate-until} evaluation protocol from Section~\ref{sec:eval}.

\begin{figure}[!ht]
    
    \centering
        \resizebox{1.0\columnwidth}{!}{
    \includegraphics[width=\linewidth]{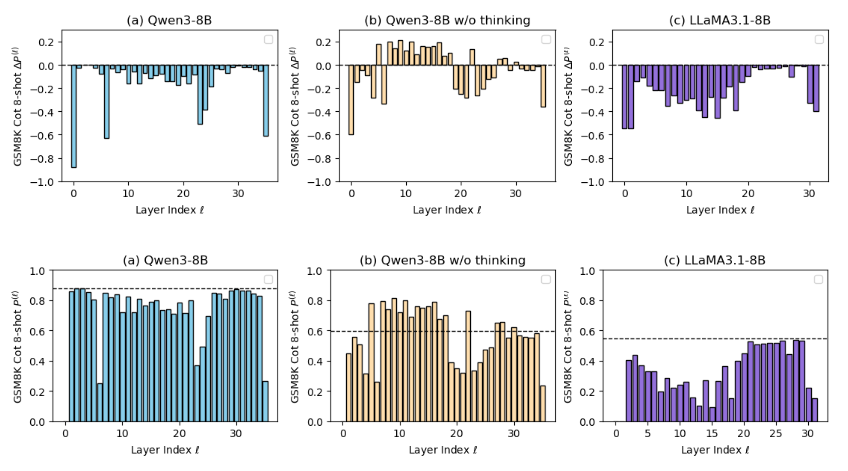}}
    \caption{Layer pruning results on the \textbf{GSM8K 8-shot} benchmark under \textbf{Qwen3-8B}, \textbf{Qwen3-8B without thinking mode}, and \textbf{LLaMA-3.1-8B}.}
    \label{fig:placeholder10}
\end{figure}

Results show that reasoning performance is highly sensitive to middle and deep layers, with ablations in these regions causing sharp drops in GSM8K accuracy. The effect is consistent across different model families (Qwen vs. LLaMA), though models with explicit CoT training (e.g., Qwen3-8B) exhibit higher baseline robustness compared to those without CoT enhancement. This confirms that multi-step mathematical reasoning relies more heavily on deeper hierarchical layers than shallow continuation tasks.

\subsection{Qwen Multi-shot Result}
    
To further examine the presence of a \textbf{reasoning layer}, we study \textbf{Qwen3-8B} under the \textbf{generate\_until} setup. We design two tasks: a \textbf{1-shot} setting with a single in-context example and a \textbf{4-shot} setting with four exemplars. For each case, we conduct layer pruning experiments on GSM8K and report absolute accuracy ($\mu$) and accuracy difference ($\Delta \mu$) based on the \textbf{generate-until} evaluation protocol.

\begin{figure}[!ht]
    
    \centering
            \resizebox{0.8\columnwidth}{!}{
    \includegraphics[width=\linewidth]{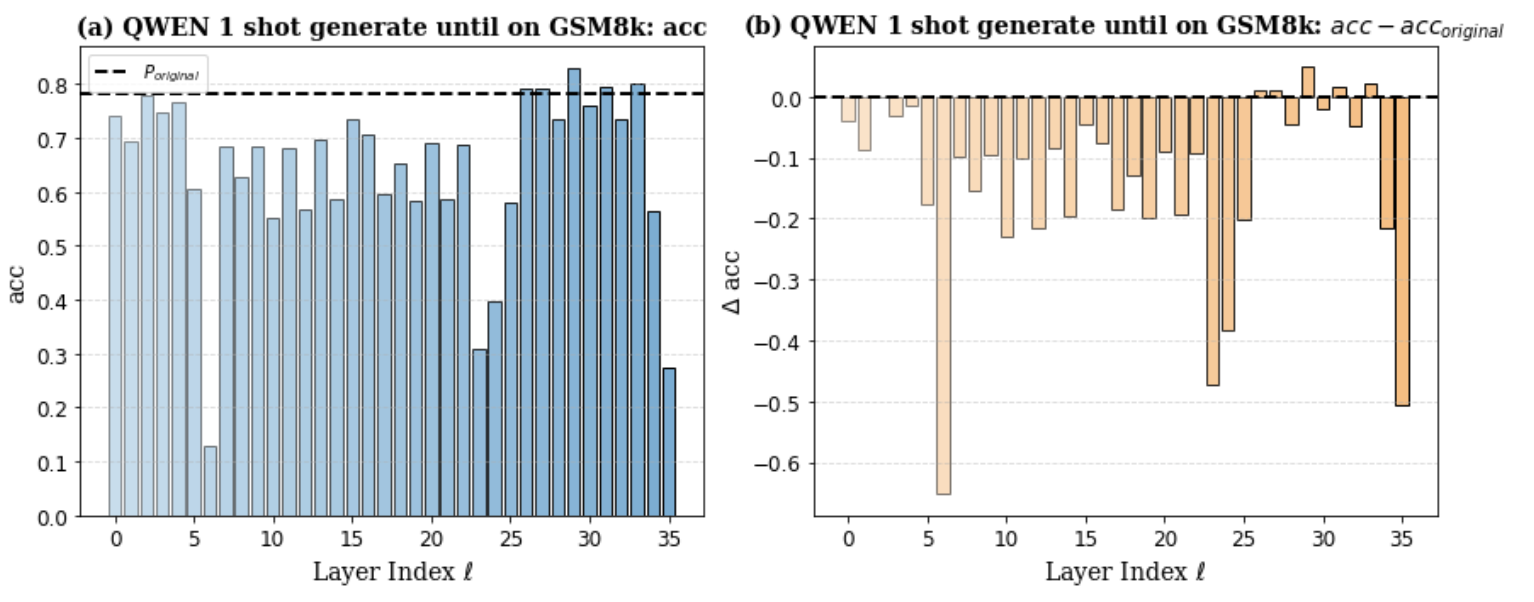}}
    \caption{Qwen3-8B on GSM8K with \textbf{1-shot generate\_until}: (a) accuracy $\mu$ (blue), (b) $\Delta \mu$ (yellow).}
    \label{fig:qwen1shot}
        
\end{figure}

\begin{figure}[!ht]
    \centering
            \resizebox{0.8\columnwidth}{!}{
    \includegraphics[width=\linewidth]{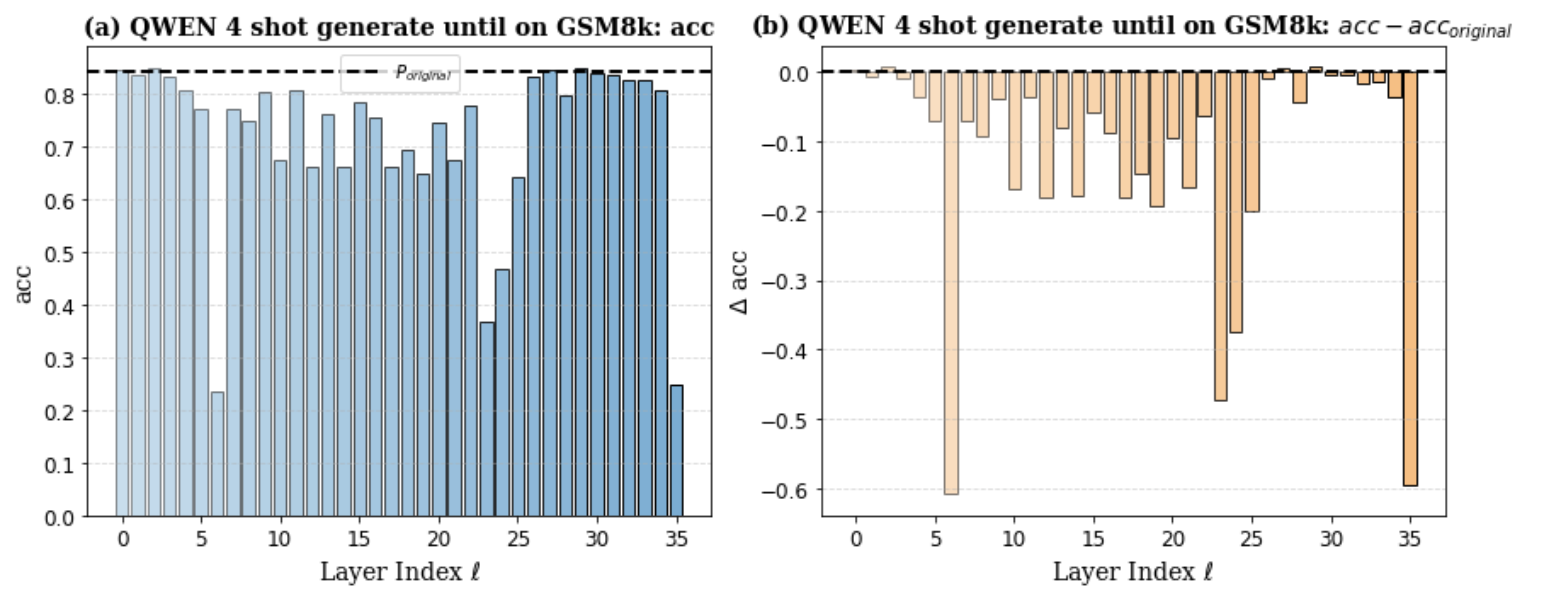}}
    \caption{Qwen3-8B on GSM8K with \textbf{4-shot generate\_until}: (a) accuracy $\mu$ (blue), (b) $\Delta \mu$ (yellow).}
    \label{fig:qwen4shot}
\end{figure}
Results in Figure~\ref{fig:qwen1shot} and Figure~\ref{fig:qwen4shot} reveal similar patterns across the two prompting strategies. In the \textbf{4-shot} setup, baseline accuracy is consistently higher than in the \textbf{1-shot} case, reflecting the benefit of additional exemplars. However, both \textbf{1-shot} and \textbf{4-shot} exhibit sharp degradations when certain layers are pruned. In particular, several intermediate layers—such as layers 6, 23, and 35—show substantial drops, with $\Delta \mu$ reaching below $-0.5$ in the most affected regions. These findings suggest that while more exemplars improve overall performance, CoT-style reasoning remains highly dependent on specific mid-to-deep layer representations.

\subsection{Qwen Reasoning Head}

Building on the layer pruning analysis, which showed that CoT-style reasoning in Qwen3-8B depends on specific intermediate and deep layers, we further investigate whether reasoning ability can be localized to individual attention heads. To this end, we conduct head pruning experiments on layer 35, one of the most critical layers identified in the GSM8K \textbf{1-shot CoT} setting in Figure~\ref{fig:qwen1shot}. For each head, we measure absolute accuracy ($\mu$) and accuracy difference ($\Delta \mu$) based on the \textbf{generate-until} evaluation protocol.

\begin{figure}[!ht]
    
    \centering
            \resizebox{0.8\columnwidth}{!}{
    \includegraphics[width=\linewidth]{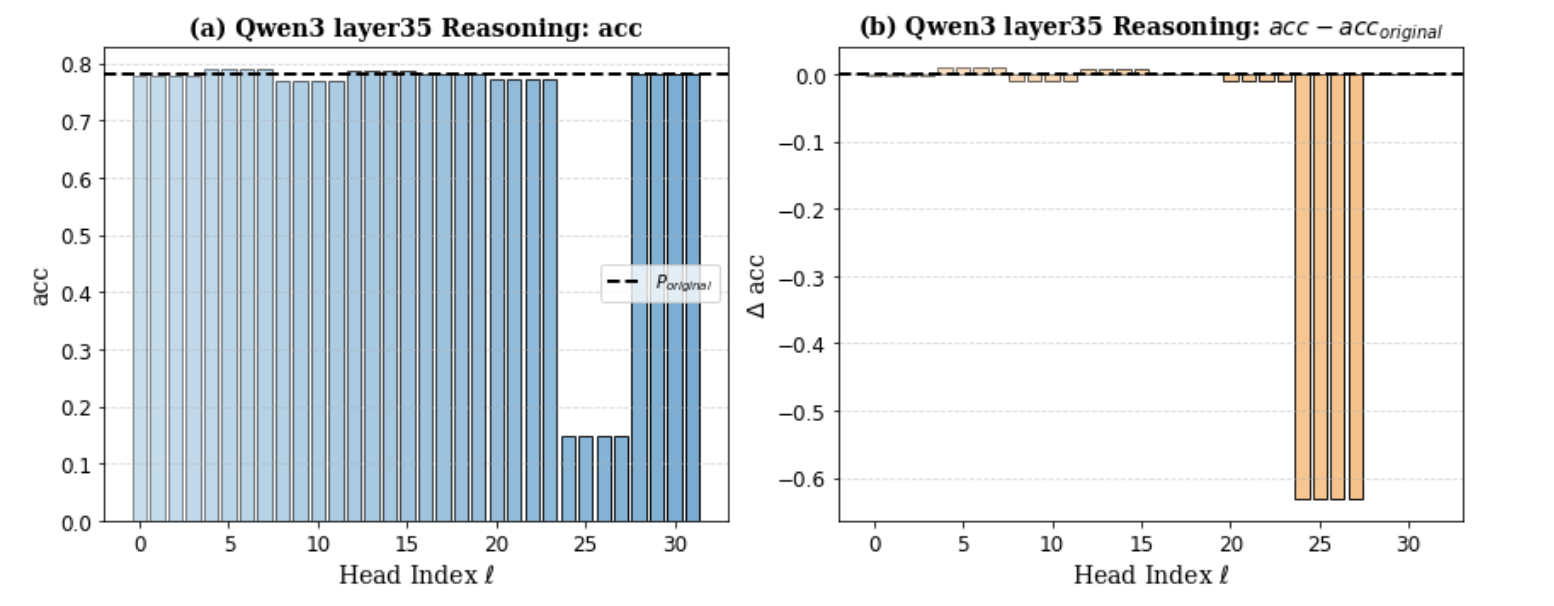}}
    \caption{Qwen3-8B on GSM8K with \textbf{Layer 35 head pruning}: (a) accuracy $\mu$ (blue), (b) $\Delta \mu$ (yellow). The x-axis denotes the head index (1–32) within layer 35.}
    \label{fig:qwen_layer35}
    
\end{figure}

As shown in Figure~\ref{fig:qwen_layer35}, reasoning ability is highly concentrated in a sparse subset of heads. Pruning certain heads at layer 35 leads to severe degradation (with $\Delta acc \approx -0.6$), while most other heads cause little to no loss when ablated. This indicates that a small number of reasoning heads dominate the completion of multi-step reasoning in the final layer. Moreover, since \textbf{Qwen3-8B} employs Grouped Query Attention~\cite{ainslie2023gqa}, the heads within the same group exhibit similar degradation patterns, further confirming that reasoning capacity is bottlenecked by specialized groups of attention heads in deep layers

\section{Distilled Reasoning Model}\label{app:distill_reasoning}
In this section, we examine whether distillation alters the depth distribution of reasoning ability. Our goal is to determine if distilled models concentrate reasoning in fewer layers or heads compared to their base counterparts.  

In addition to \textbf{Qwen}, we also evaluate a distilled \textbf{LLaMA-3.1-8B} on GSM8K to probe the location of reasoning layers.

\subsection{LLaMA-3.1-8B (distilled) on GSM8K under CoT}
Distilled models retain strong reasoning ability. To locate where this ability resides in depth, we analyze \textbf{LLaMA-3.1-8B} and its distilled variant Deepseek model~\cite{deepseekai2025deepseekr1} on the \textbf{GSM8K} benchmark using chain-of-thought (CoT) prompting. We perform layer pruning and measure raw accuracy (\textbf{acc}) and accuracy difference relative to the model ($\Delta \mu$) as functions of the layer index.

\begin{figure}[!ht]
    \centering
    \includegraphics[width=0.8\linewidth]{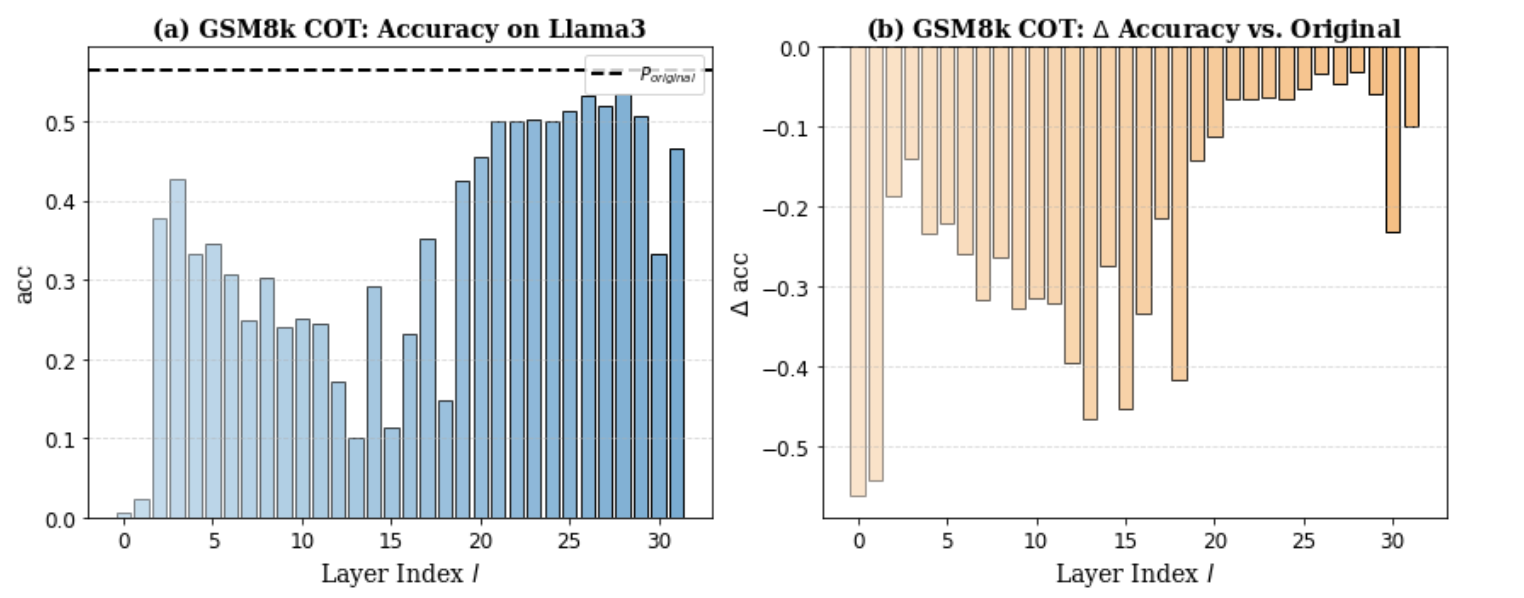}
    \caption{\textbf{LLaMA-3.1-8B} on GSM8K with \textbf{CoT prompting}: (a) accuracy, (b) $\Delta \textbf{acc}$, as functions of layer index.}
    \label{fig:llama_cot}
\end{figure}

\begin{figure}[!ht]
    \centering
    \includegraphics[width=0.8\linewidth]{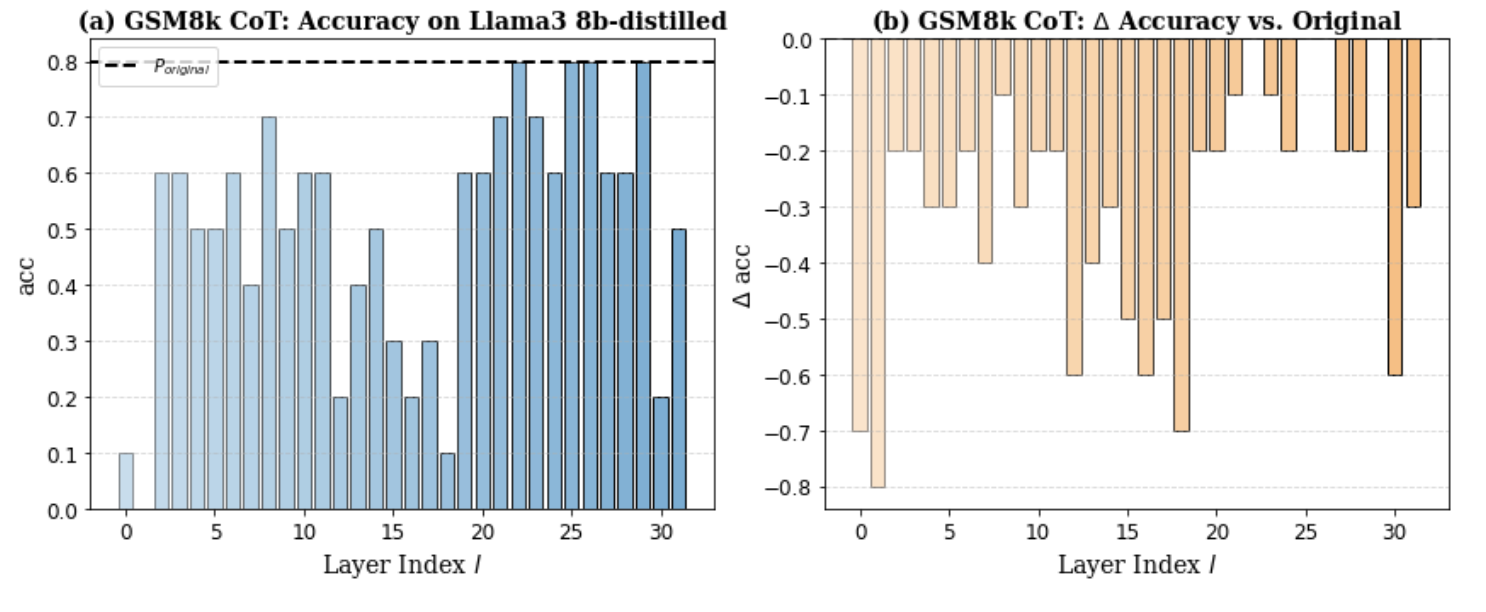}
    \caption{\textbf{LLaMA-3.1-8B (distilled)} on GSM8K with \textbf{CoT prompting}: (a) accuracy, (b) $\Delta \textbf{acc}$, as functions of layer index.}
    \label{fig:llama_dist_cot}
\end{figure}

Figure~\ref{fig:llama_cot} shows the base model under CoT, and Figure~\ref{fig:llama_dist_cot} shows the distilled model under the same protocol. Across both models, reasoning performance is highly sensitive to shallow and middle layers, with $\Delta \mu$ drops reaching $-0.6$ or lower in the most affected regions. The distilled variant yields a higher baseline accuracy and slightly improved robustness in deeper layers, but it still exhibits notable vulnerability when early and mid-depth layers are ablated. These results indicate that, even after distillation, CoT-style reasoning depends on representations that form primarily in the shallow-to-mid depth range, with deeper layers contributing to stabilization rather than fully offsetting losses from earlier layers.

\subsection{LLaMA-distilled Reasoning Head}
Building on the layer pruning analysis in Fugure~\ref{fig:llama_dist_cot}, which revealed that reasoning in LLaMA-3.1-8B-distilled emerges in specific middle and deep layers, we further investigate whether this functionality can be localized to individual attention heads. To this end, we conduct head pruning experiments at three representative layers: layer 12 (early–middle depth), layer 18 (middle depth), and layer 30 (final layer). For each case, we measure absolute accuracy ($\mu$) and accuracy difference relative to the unablated model ($\Delta \mu$).

\begin{figure}[!ht]
    \centering
    \includegraphics[width=0.8\linewidth]{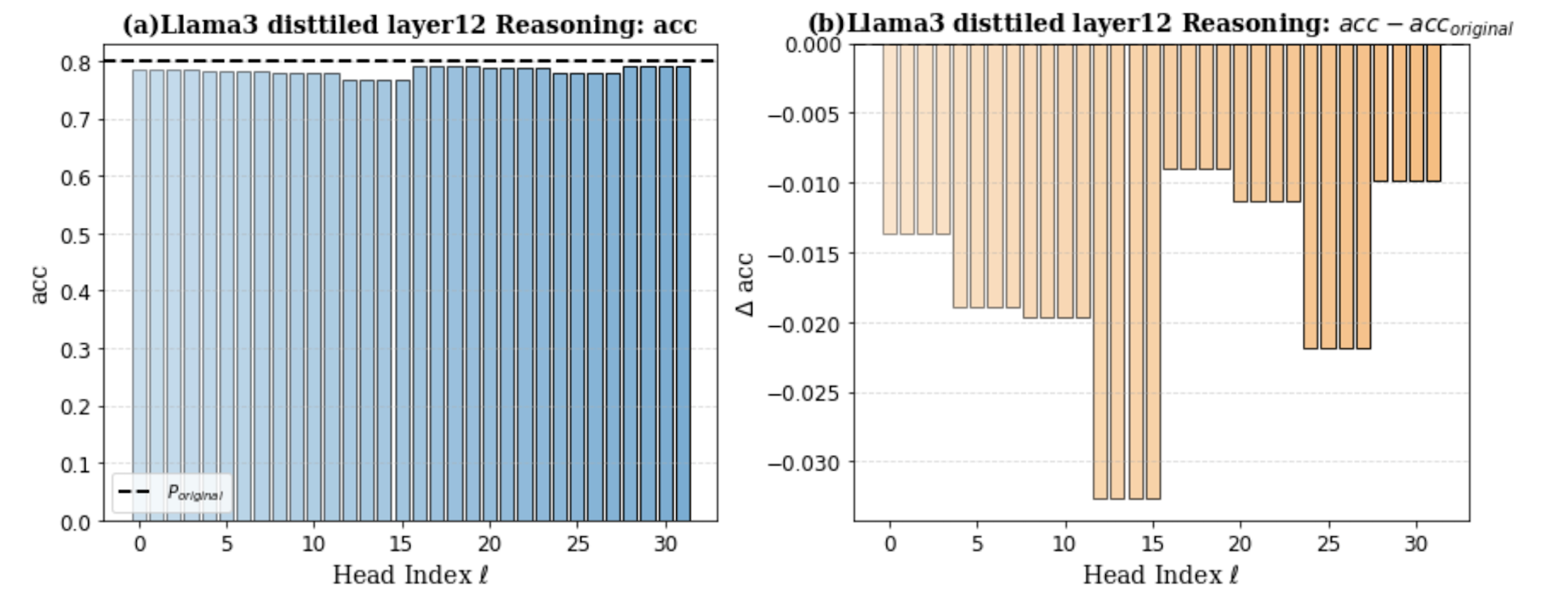}
    \caption{LLaMA-3.1-8B-distilled on GSM8K with \textbf{Layer 12 head pruning}: (a) accuracy $\mu$ per head, (b) $\Delta \mu$ per head.}
    \label{fig:llama_dist_layer12}
\end{figure}

\begin{figure}[!ht]
    \centering
    \includegraphics[width=0.8\linewidth]{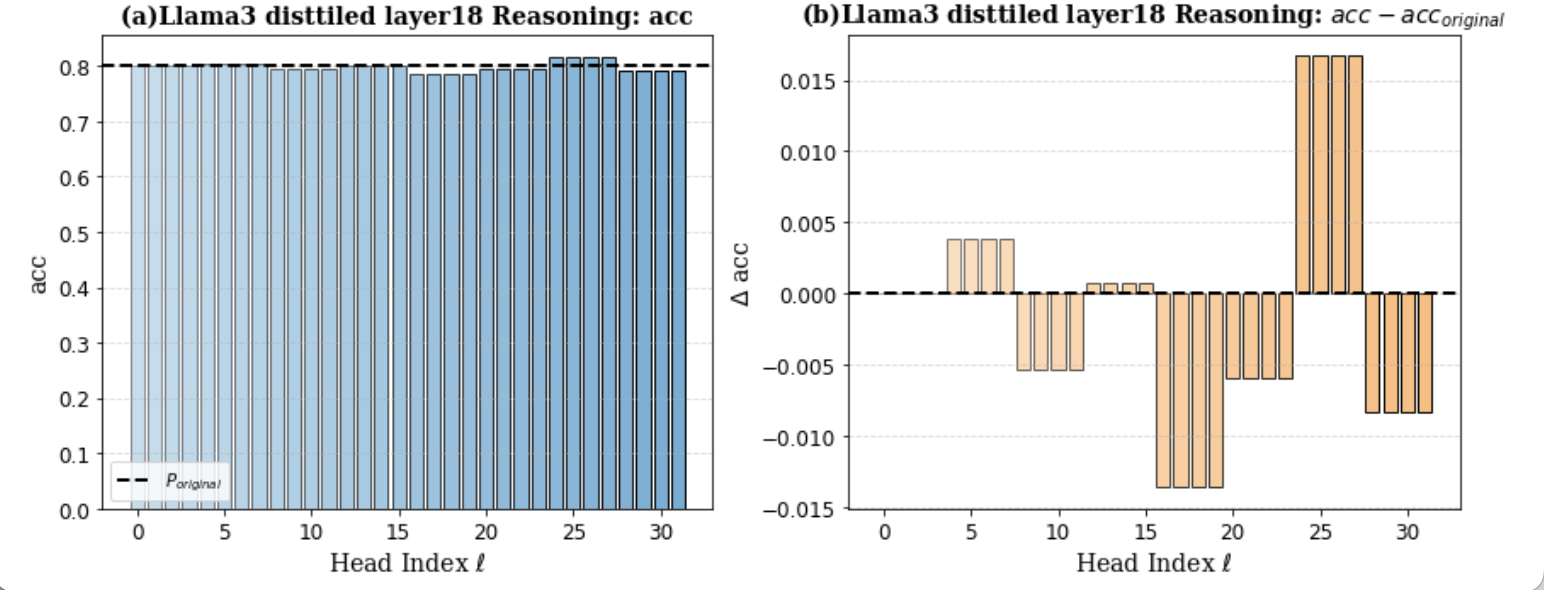}
    \caption{LLaMA-3.1-8B-distilled on GSM8K with \textbf{Layer 18 head pruning}: (a) accuracy $\mu$ per head, (b) $\Delta \mu$ per head.}
    \label{fig:llama_dist_layer18}
\end{figure}

\begin{figure}[!ht]
    \centering
    \includegraphics[width=0.8\linewidth]{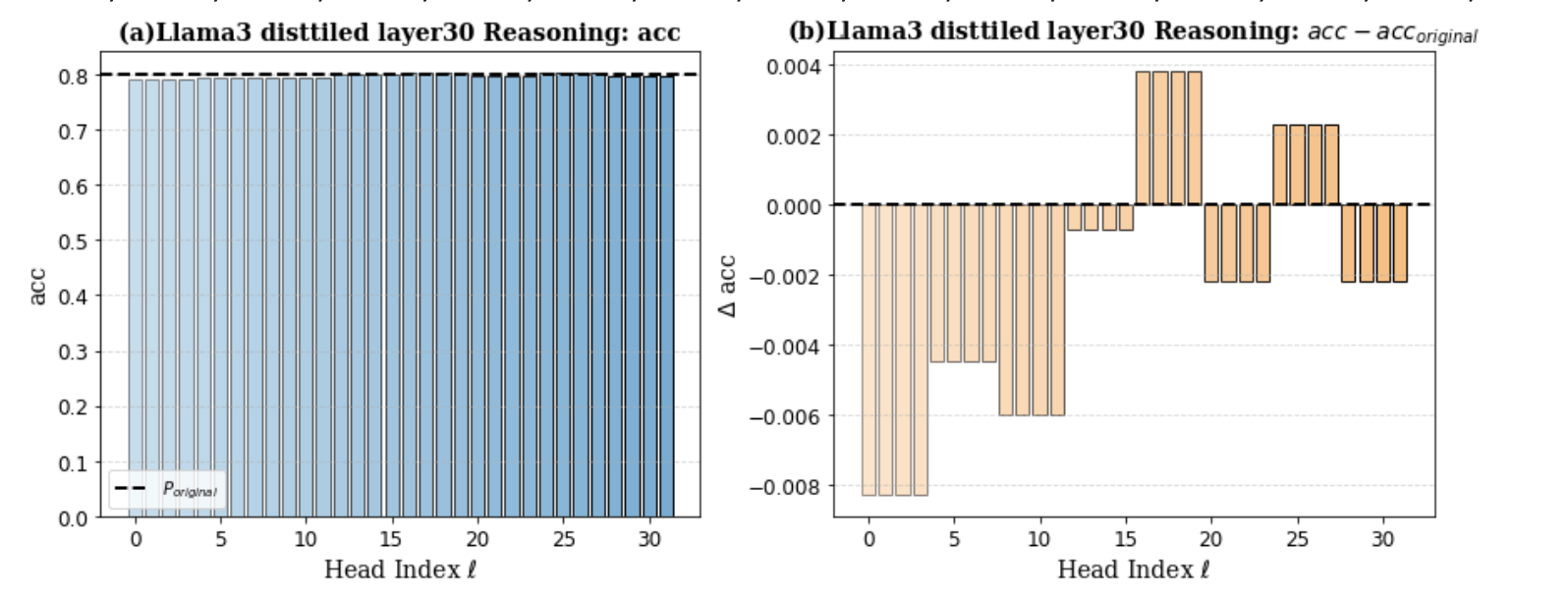}
    \caption{LLaMA-3.1-8B-distilled on GSM8K with \textbf{Layer 30 head pruning}: (a) accuracy per head, (b) $\Delta \mu$ per head.}
    \label{fig:llama_dist_layer30}
\end{figure}

The results show that reasoning ability in the distilled model is indeed localized to a sparse subset of heads across different depths. At layer 12, pruning certain heads leads to moderate degradation ($\Delta \mu \approx -0.03$), revealing early reasoning-sensitive heads. At layer 18, we observe mixed effects: some heads contribute positively, slightly boosting performance when pruned, while others cause drops ($\Delta \mu \approx -0.015$). At layer 30, pruning specific deep heads reduces performance ($\Delta \mu \approx -0.008$), suggesting that reasoning is consolidated by a few specialized heads in the final layer. Overall, these findings indicate that distilled reasoning models distribute reasoning functionality across multiple layers but still rely heavily on a small number of specialized attention heads.

\section{Delta Model} 
In this section, we investigate how distillation effects can be isolated and analyzed through delta model replacement. Our goal is to determine whether improvements from distillation are concentrated in specific layers, and how replacing projections between models influences reasoning robustness.  

\subsection{Delta Model: Distillation-Base}

We evaluate the effect of pruning ($\Delta W$) combined with output projection $W$ on the \textbf{GSM8K Chain-of-Thought (CoT)} benchmark, comparing the \textbf{DeepSeek-LLaMA3-distilled} model against the original \textbf{LLaMA-3.1} under layer pruning. The goal is to examine whether certain layers are crucial for distillation robustness. 

Specifically, we adopt a delta model setting, where the output projection matrix of a given layer in one model is replaced by that of another model. Formally, let $W^{(l)}_{\text{src}}$ denote the output projection at layer $l$ of the source model (e.g., DeepSeek-LLaMA3-distilled), and $W^{(l)}_{\text{tgt}}$ the corresponding matrix of the target model (e.g., LLaMA-3.1). The delta replacement is defined as:
\begin{equation}
W^{(l)}_{\text{tgt}}   \longleftarrow   W^{(l)}_{\text{tgt}} + \Delta W^{(l)}, 
\quad \text{where} \quad 
\Delta W^{(l)} = W^{(l)}_{\text{src}} - W^{(l)}_{\text{tgt}}.
\end{equation}

This operation effectively injects the representation learned by one model into the corresponding layer of the other, allowing us to isolate whether improvements from distillation are concentrated in specific layers.

\begin{figure}[!ht]
    \centering
    \includegraphics[width=0.8\linewidth]{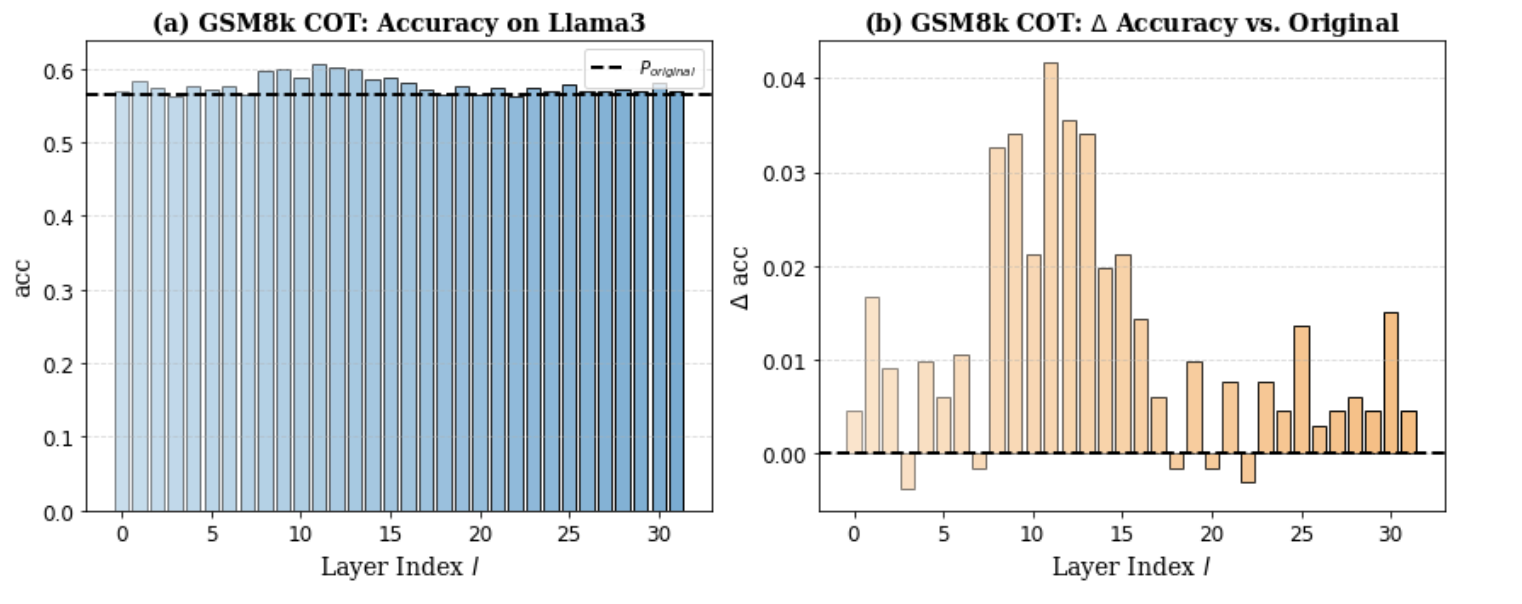}
    \caption{layer pruning results of \textbf{DeepSeek-LLaMA3-distilled} on GSM8k CoT. Left: absolute accuracy  ($\mu$). Right: relative accuracy difference ($\Delta \mu$) compared to the unablated model.}
    \label{fig:deepseek_cot}
\end{figure}

For both models, we report (i) absolute accuracy and (ii) accuracy difference relative to the unablated model across all layers (Figure~\ref{fig:deepseek_cot}).

Results show that the \textbf{DeepSeek-distilled model} maintains high accuracy across almost all layers, with only mild fluctuations in $\Delta \mu$ and even small improvements in certain middle layers (Figure~\ref{fig:deepseek_cot}). When layers of \textbf{LLaMA-3.1} are replaced with those from the distilled model, we observe slight gains in the middle layers, suggesting that distilled representations at intermediate depths help enhance the reasoning ability of the base LLaMA model.

\subsection{Delta Model: Reverse Replacement}

\begin{figure}[!ht]
    \centering
    \includegraphics[width=0.8\linewidth]{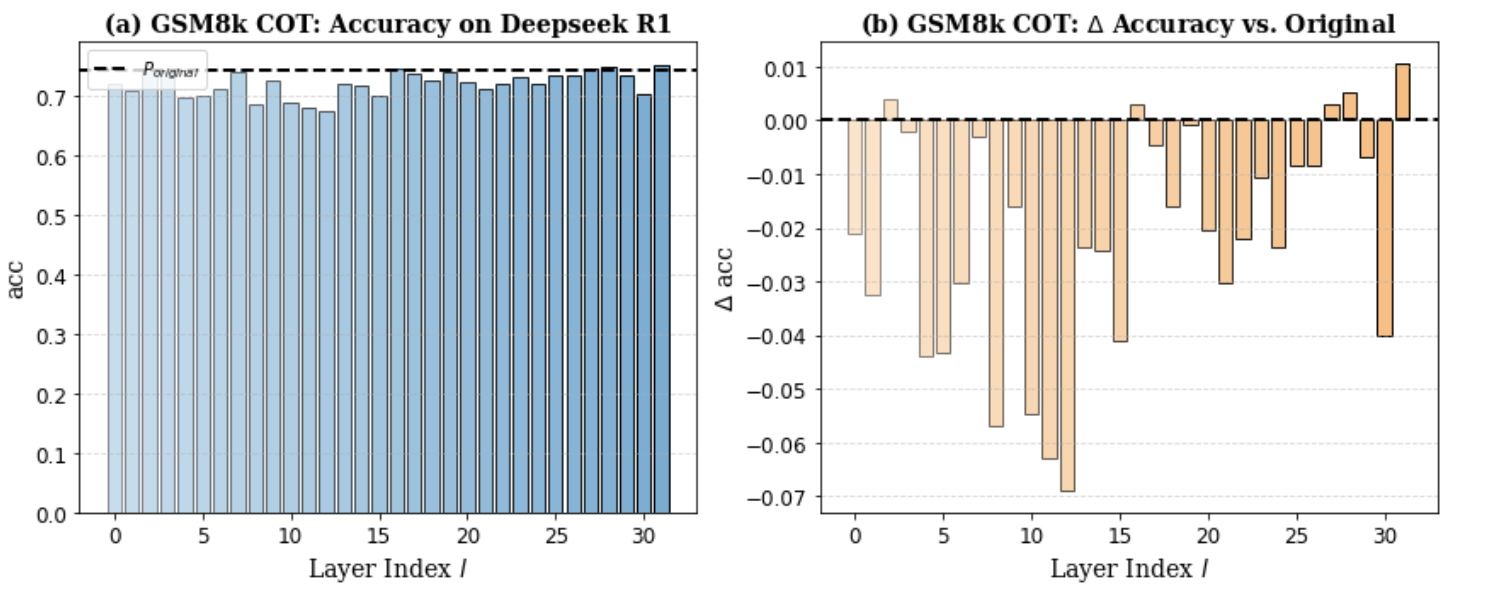}
    \caption{Reverse replacement on \textbf{LLaMA-3.1 distilled} for GSM8K CoT: for each layer $l$, the distilled layer’s output projection $W_o^{(l)}$ is replaced by the corresponding \textbf{LLaMA-3.1-8B} layer. Left: absolute accuracy ($\mu$). Right: accuracy difference ($\Delta \mu$) relative to the distilled baseline. Lower $\Delta \mu$ indicates performance loss after substituting a distilled layer with a base layer.}
    \label{fig:deepseek_r1_delta}
\end{figure}

We further evaluate \textbf{pruning ($\Delta W$) combined with output projection $W_o$} under a \textbf{reverse replacement} setting, in which we progressively substitute each layer of the \textbf{LLaMA-3.1 distilled} model with the corresponding layer from the original \textbf{LLaMA-3.1-8B} (i.e., for layer $l$, $W^{(l)}_{o,\text{distilled}} \leftarrow W^{(l)}_{o,\text{base}}$). We run this procedure on GSM8K CoT and report absolute accuracy ($\mu$) and accuracy difference ($\Delta \mu$) relative to the distilled baseline. The results are presented in Figure~\ref{fig:deepseek_r1_delta}.

Results show that \textbf{LLaMA-3.1 distilled} suffer performance degradation once its layers are replaced by those of the original \textbf{LLaMA-3.1}. In Figure~\ref{fig:deepseek_r1_delta}, early and middle layers exhibit negative $\Delta \mu$ (up to $-0.06$), while later layers remain comparatively stable, indicating that replacing distilled representations with base LLaMA layers weakens reasoning robustness. These results demonstrate that distilled models encode reasoning capacity in a more robust layer distribution, and that reverse replacement with base model layers diminishes this advantage, confirming that distillation plays a key role in strengthening reasoning resilience against pruning.

\subsection{Accumulative Layer Replacement Analysis}

\begin{figure}[!ht]
    \centering
    \includegraphics[width=0.8\linewidth]{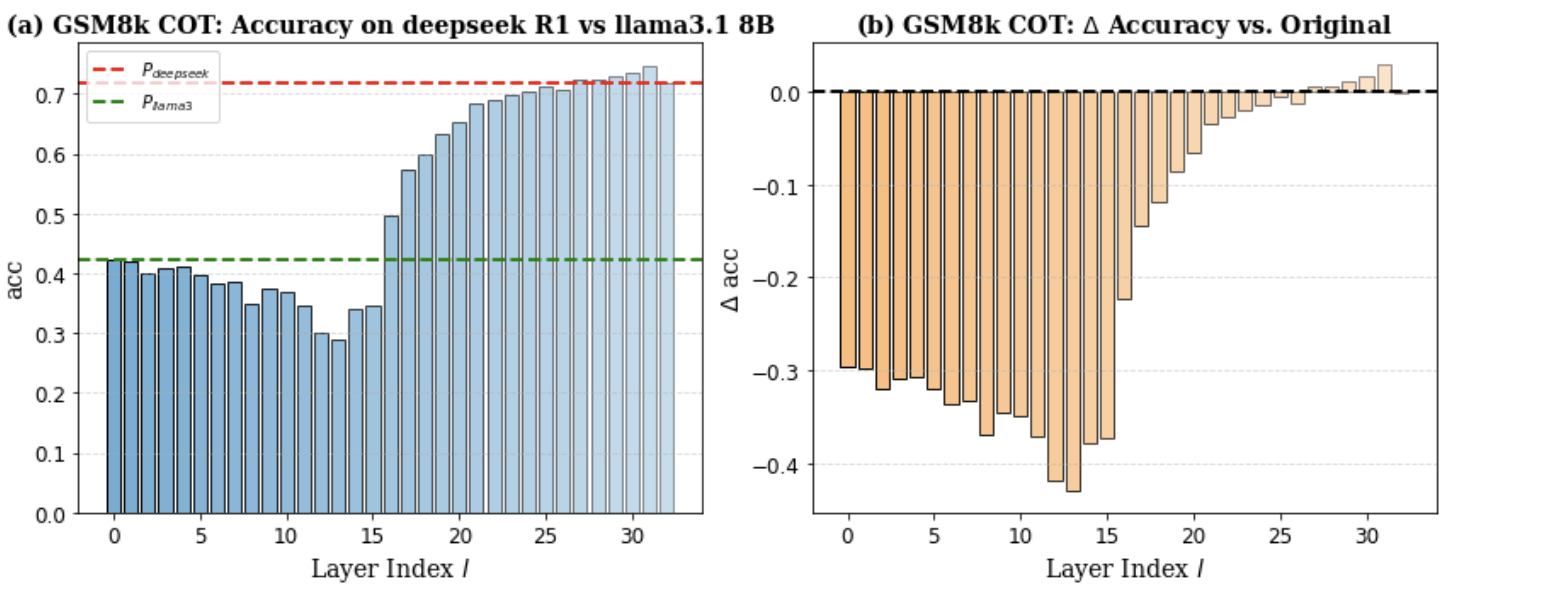}
    \caption{Accumulative layer replacement: \textbf{DeepSeek R1 replaced by LLaMA-3.1-8B} on GSM8k CoT. Left: absolute accuracy ($\mu$). Right: relative accuracy difference ($\Delta \mu$) compared to the original DeepSeek R1 baseline.}
    \label{fig:deepseek_to_llama}
\end{figure}

\begin{figure}[!ht]
    \centering
    \includegraphics[width=0.8\linewidth]{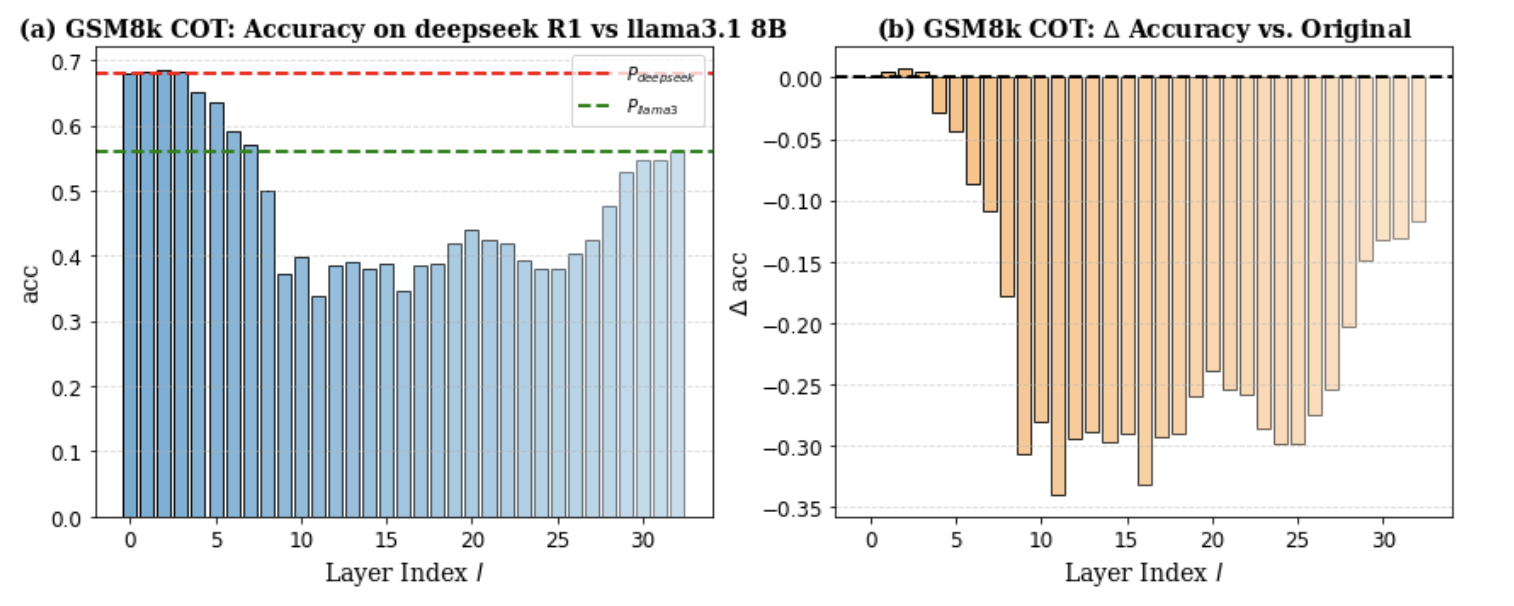}
    \caption{Accumulative layer replacement: \textbf{LLaMA-3.1-8B replaced by DeepSeek R1} on GSM8k CoT. Left: absolute accuracy ($\mu$). Right: relative accuracy difference ($\Delta \mu$) compared to the original LLaMA-3.1-8B baseline.}
    \label{fig:llama_to_deepseek}
\end{figure}

We further investigate the effect of \textbf{accumulative layer replacement} under pruning ($\Delta W$) combined with output projection $W_o$ on the \textbf{GSM8K Chain-of-Thought (CoT)} benchmark. In the first setting, we progressively replace layers of \textbf{DeepSeek R1} with those of \textbf{LLaMA-3.1-8B} (Figure~\ref{fig:deepseek_to_llama}), while in the second setting we reverse the process, gradually substituting layers of \textbf{LLaMA-3.1-8B} with those of \textbf{DeepSeek R1} (Figure~\ref{fig:llama_to_deepseek}).

Results show that in the \textbf{DeepSeek $\rightarrow$ LLaMA replacement} setting (Figure~\ref{fig:deepseek_to_llama}), accuracy initially drops sharply when the first few layers are replaced (up to $-0.4$), suggesting that shallow layers are critical for preserving DeepSeek’s distilled reasoning ability. However, as more layers are replaced, performance gradually converges to the LLaMA baseline. Conversely, in the \textbf{LLaMA $\rightarrow$ DeepSeek replacement} setting (Figure~\ref{fig:llama_to_deepseek}), we observe steady accuracy gains as more DeepSeek layers accumulate, indicating that distilled DeepSeek layers introduce robustness and reasoning improvements even when partially integrated.  

Overall, these experiments highlight that early and middle layers are crucial for transferring distilled knowledge, while later layers can be replaced with relatively smaller impact. This supports the view that distillation redistributes reasoning capacity across the depth of the network but still depends critically on shallow-to-mid representations.

\section{Conclusion}
\label{sec:conclusion}

We systematically analyzed depth usage in large language models via layer pruning across tasks, metrics, and model families. Results show that layer contributions are highly uneven: shallow layers dominate likelihood and retrieval, while mid-to-deep layers are essential for reasoning and generation. Taken together, depth usage is inherently task-dependent, highly metric-sensitive, and strongly model-specific, highlighting the critical need for task-aware evaluation and offering practical guidance for compression and future model design.

\bibliographystyle{plainnat}
\bibliography{custom}

\end{document}